\documentclass[letterpaper, 10 pt, conference]{ieeeconf}  

\usepackage[top=54pt, bottom=54pt, left=0.75in, right=0.75in]{geometry}

\IEEEoverridecommandlockouts                              

\usepackage{tabularx}
\usepackage{array}

\usepackage{float} 

\usepackage{amssymb}
\usepackage{newtxtext,newtxmath} 
\usepackage{graphicx}
\usepackage[export]{adjustbox}
\usepackage{url}
\usepackage{latexsym}
\usepackage{algpseudocode}
\usepackage{amsmath}
\usepackage{lipsum}
\usepackage{caption}
\usepackage{bbding}
\usepackage{wasysym}
\usepackage{blindtext}
\usepackage{booktabs}
\usepackage{tabularx}
\usepackage{array}
\usepackage{pifont}
\usepackage[export]{adjustbox}
\usepackage{float}
\usepackage{multirow}

\usepackage{amsthm}
\usepackage{xcolor}
\usepackage{cite}

\title{\LARGE \bf
A Survey of Reinforcement Learning for Optimization in Automation
}

\author{Ahmad Farooq*$^{1}$ and Kamran Iqbal$^{2}$
\thanks{This work is not supported by any organization. This work is a preprint version of the paper published in the 2024 IEEE 20th International Conference on Automation Science and Engineering (CASE) held from
August 28 to September 1, 2024, in Bari, Italy. The final version is available at IEEE Xplore under the conference proceedings.}
\thanks{*Corresponding Author: Ahmad Farooq}
}

\begin{document}

\maketitle
\thispagestyle{empty}
\pagestyle{empty}

\begin{abstract}
Reinforcement Learning (RL) has become a critical tool for optimization challenges within automation, leading to significant advancements in several areas. This review article examines the current landscape of RL within automation, with a particular focus on its roles in manufacturing, energy systems, and robotics. It discusses state-of-the-art methods, major challenges, and upcoming avenues of research within each sector, highlighting RL's capacity to solve intricate optimization challenges. The paper reviews the advantages and constraints of RL-driven optimization methods in automation. It points out prevalent challenges encountered in RL optimization, including issues related to sample efficiency and scalability; safety and robustness; interpretability and trustworthiness; transfer learning and meta-learning; and real-world deployment and integration. It further explores prospective strategies and future research pathways to navigate these challenges. Additionally, the survey includes a comprehensive list of relevant research papers, making it an indispensable guide for scholars and practitioners keen on exploring this domain.


Manufacturing, Energy Systems, Robotics
\end{abstract}

\section{INTRODUCTION}
\subsection{Motivation}
Reinforcement learning (RL) has emerged as a effective framework for sequential decision-making problems, enabling agents to learn optimal policies through interaction with the environment \cite{sutton2018reinforcement, mnih2015human}
In recent years, RL has achieved remarkable success in various domains, including manufacturing \cite{li2023deep}, energy systems \cite{perera2021applications}, and robotics \cite{kober2013reinforcement}. The key advantage of RL lies in its ability to learn from trial-and-error experience without requiring explicit supervision or a predefined model.

Simultaneously, optimization problems are ubiquitous in automation, spanning diverse areas such as production scheduling \cite{esteso2023reinforcement}, process control \cite{nian2020review}, and inventory management \cite{boute2022deep}. These problems often involve complex decision-making under uncertainty, large-scale combinatorial search spaces, and dynamic environments. Traditional optimization approaches, such as mathematical programming and metaheuristics, have been extensively studied and applied to automation problems \cite{blum2003metaheuristics}.
However, they often struggle with scalability, adaptability, and the need for domain-specific knowledge.

The intersection of RL and optimization in automation presents a promising avenue for addressing these challenges. By leveraging the power of RL to learn from experience and adapt to changing conditions, we can develop more efficient, flexible, and robust optimization algorithms for automation tasks
\cite{li2017deep, arulkumaran2017deep}. This has led to a growing body of research on RL-based optimization in various automation domains, which is the focus of this survey.

\subsection{Scope and Contributions}
This survey paper aims to provide a comprehensive overview of RL techniques for optimization in automation. We focus on three key application domains: manufacturing, energy systems, and robotics. In each domain, we review representative works that demonstrate the effectiveness of RL in solving optimization problems and discuss the unique challenges and opportunities.

The main contributions of this survey are as follows:

1. We provide a systematic categorization of RL-based optimization approaches in automation, highlighting their strengths and limitations.

2. We discuss the state-of-the-art RL algorithms used for optimization in each application domain.

3. We identify common challenges faced by RL-based optimization in automation, including sample efficiency and scalability; safety and robustness; interpretability and trustworthiness; transfer learning and meta-learning; and real-world deployment and integration, and discuss potential solutions and future research directions.

4. We present a comprehensive bibliography of relevant research papers, serving as a valuable resource for researchers and practitioners interested in this field.

To the best of our knowledge, this is the first survey paper that specifically focuses on RL for optimization in automation, covering a wide range of application domains and providing insights into the current state and future prospects of this rapidly growing field.

\subsection{Organization of the Paper}
The remainder of this survey is organized as follows: Section II focuses on the applications of RL-based optimization in three major domains: manufacturing, energy systems, and robotics. For each domain, we provide a comparative analysis of the selected papers, highlighting their key findings, methodologies, and contributions. We also discuss the domain-specific challenges and opportunities. Section III discusses the common challenges faced by RL-based optimization in automation. We present an overview of the potential solutions and future research directions to address these challenges. Finally, Section IV concludes the survey, summarizing the key takeaways. 

\begin{figure*}[ht]
    \centering
    \includegraphics[width=0.67\linewidth]{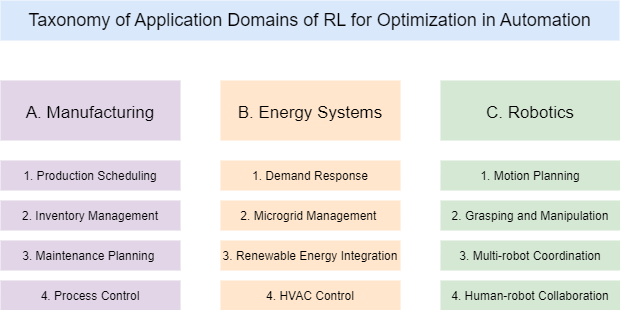}
    \caption{Taxonomy of Application Domains of  RL for Optimization in Automation}
    \label{fig:Taxonomy_App_Dom.png}
\end{figure*}

\section{APPLICATION DOMAINS}
Reinforcement Learning (RL) has revolutionized automation in Manufacturing, Energy Systems, and Robotics. Figure \ref{fig:Taxonomy_App_Dom.png} shows these major domains and their sub-domains that we will discuss in this section.
\subsection{Manufacturing}
RL is revolutionizing manufacturing through advancements in production scheduling, inventory management, maintenance planning, and process control, showcasing its potential to tackle complex optimization challenges within this sector. In production scheduling, RL methods surpass traditional models by adeptly handling uncertainties, thereby enhancing profitability and customer service \cite{hubbs2020deep, shi2020intelligent, guo2020reinforcement, esteso2023reinforcement, mowbray2022distributional}. For inventory management, RL techniques, particularly Deep Reinforcement Learning (DRL) and Multi-agent Reinforcement Learning (MARL), offer innovative solutions for managing stochastic demands and complex supply chains, leading to improved sales and reduced wastage \cite{boute2022deep, sultana2020reinforcement, de2022reward, khirwar2023cooperative, leluc2023marlim}. Maintenance planning benefits from RL's dynamic optimization capabilities, utilizing real-time data for maintenance schedules, thus improving system reliability and reducing downtimes \cite{ogunfowora2023reinforcement, yousefi2020reinforcement, yousefi2022dynamic, andrade2021aircraft, thomas2021network}. In process control, RL's adaptability ensures product quality and operational efficiency, with methodologies like Explainable RL and DRL enhancing process understanding and control strategies \cite{viharos2021reinforcement, nian2020review, kuhnle2022explainable, mowbray2021using, li2023reinforcement}. Future directions point towards developing risk-sensitive formulations, leveraging real-world data, and integrating smart systems to further enhance manufacturing efficiency. Table \ref{tab:manufacturing} encapsulates these insights by outlining key objectives, challenges addressed, RL approaches, outcomes, and future directions, alongside representative studies that underscore RL's transformative impact on manufacturing.

\newcolumntype{Y}{>{\centering\arraybackslash}X}
\newcommand{\cmark}{\ding{51}}
\newcommand{\xmark}{\ding{55}}

\renewcommand{\arraystretch}{1} 
\renewcommand\tabularxcolumn[1]{m{#1}} 

\begin{table*}
\centering
\begin{tabularx}{\textwidth}{|Y|Y|Y|Y|Y|}
\hline

\textbf{Feature/Criteria} & \textbf{Production Scheduling} & \textbf{Inventory Management} & \textbf{Maintenance Planning} & \textbf{Process Control} \\

\hline
\textbf{Key Objectives} & Optimize allocation of tasks to resources over time & Balance supply with demand, minimize costs, and ensure timely product availability & Minimize downtime, extend asset life, ensure safety & Ensure product quality, operational efficiency, and safety \\
\hline
\textbf{Challenges Addressed} & Handling complexities and uncertainties in scheduling tasks & Stochastic demand, perishable goods, multi-echelon supply chains & Dynamic maintenance planning under system degradation & Controlling complex manufacturing processes \\
\hline
\textbf{RL Approaches} & DQN \cite{esteso2023reinforcement}, Distributional RL \cite{mowbray2022distributional}, DRL \cite{shi2020intelligent}, A2C \cite{hubbs2020deep} & DQN \cite{de2022reward}, PPO \cite{leluc2023marlim}, A2C \cite{sultana2020reinforcement}, DRL \cite{boute2022deep}, Cooperative MARL \cite{khirwar2023cooperative}

& 
Multi-Agent Actor Critic \cite{thomas2021network}, Deep Q-learning \cite{andrade2021aircraft, yousefi2022dynamic}, Q-learning \cite{yousefi2020reinforcement}

& TRPO \cite{kuhnle2022explainable}, DDPG \cite{nian2020review}, Dynamic Q-table \cite{viharos2021reinforcement} \\
\hline
\textbf{Methodology Highlights} & Superiority over traditional mixed integer linear programming models, competitive performance against heuristic methods & Comprehensive roadmap for DRL deployment, novel frameworks for multi-agent hierarchical inventory management & Analysis of RL/DRL applications, dynamic maintenance policies using Q-learning & Adaptation of RL for Statistical Process Control (SPC), integration of domain expertise, apprenticeship learning \\
\hline
\textbf{Outcomes} & Increased profitability, reduced inventory levels, improved customer service & Maximized sales, minimized perishable product wastage, optimized supply chain needs & Reduced maintenance activities, enhanced fleet availability, adapted maintenance policies & Enhanced SPC adaptability, improved control policies, handling nonlinearities in manufacturing \\
\hline
\textbf{Future Directions} & Development of risk-sensitive formulations, leveraging real-world data & Advanced cooperative strategies among agents, leveraging custom GPU-parallelized environments & Integration with smart factory systems, leveraging condition monitoring data & Utilization of real-world data for training, improving RL training efficiency \\
\hline
\textbf{Representative Studies} & 
Hubbs et al. \cite{hubbs2020deep}, Shi et al. \cite{shi2020intelligent}, Guo et al. \cite{guo2020reinforcement}, Esteso et al. \cite{esteso2023reinforcement}, Mowbray et al. \cite{mowbray2022distributional} 

& 
Boute et al. \cite{boute2022deep}, Sultana et al. \cite{sultana2020reinforcement}, De Moor et al. \cite{de2022reward}, Khirwar et al. \cite{khirwar2023cooperative}, Leluc et al. \cite{leluc2023marlim}

& 

Ogunfowora and Najjaran \cite{ogunfowora2023reinforcement}, Yousefi et al. \cite{yousefi2020reinforcement}, Yousefi et al. \cite{yousefi2022dynamic}, Andrade et al. \cite{andrade2021aircraft}, Thomas et al. \cite{thomas2021network}

& 
Viharos and Jakab \cite{viharos2021reinforcement}, Nian, Liu, and Huang \cite{nian2020review}, Kuhnle et al. \cite{kuhnle2022explainable}, Mowbray et al. \cite{mowbray2021using}, and Li, Du, Jiang \cite{li2023reinforcement}

\\
\hline
\end{tabularx}
\caption{Comparison of RL Approaches for Optimization in Manufacturing}
\label{tab:manufacturing}
\end{table*}

\subsection{Energy Systems}
RL and DRL are transforming energy systems, offering innovative solutions across demand response, microgrid management, renewable energy integration, and Heating, Ventilation, and Air Conditioning (HVAC) control to optimize and enhance grid stability, sustainability, and energy efficiency. Demand response strategies benefit from DRL and MARL to dynamically adjust energy usage in response to utility signals, achieving up to 22\% energy savings and more efficient electricity management \cite{azuatalam2020reinforcement, jang2021using, ahrarinouri2020multiagent, lu2021deep, lu2020multi, zhang2021testbed}. In microgrid management, DRL and MARL approaches enhance grid resilience by optimizing energy distribution and usage, resulting in improved cost efficiency and increased system reliability \cite{nakabi2021deep, hu2021energy, zhang2023deep, zhang2022energy, shojaeighadikolaei2021weather, du2019intelligent}. For renewable energy integration, RL's capability to handle the variability of renewable sources leads to more effective energy dispatch strategies, ensuring grid stability and maximizing the use of renewable resources \cite{yang2020reinforcement, cao2020reinforcement, chen2022reinforcement, sivamayil2023systematic, perera2021applications, ahrarinouri2020multiagent}. HVAC systems, as major energy consumers, see optimizations through DRL and batch RL methods, achieving significant reductions in energy consumption while maintaining occupant comfort \cite{azuatalam2020reinforcement, zhong2022end, sierla2022review, liu2022safe, yuan2021study, biemann2021experimental}. Looking ahead, the future promises advancements in adaptive strategies and bridging the simulation-experiment gap for demand response, enhanced learning efficiency for microgrid management, scalability and adaptability improvements in renewable energy integration, and wider applicability of pre-trained models for HVAC control. This narrative is encapsulated in the Table \ref{tab:energy}, which outlines the key objectives, challenges addressed, RL methodologies, and outcomes for each subdomain, alongside future research directions and representative studies illustrating RL's significant role in advancing energy systems.

\newcolumntype{Y}{>{\centering\arraybackslash}X}

\renewcommand{\arraystretch}{1} 
\renewcommand\tabularxcolumn[1]{m{#1}} 

\begin{table*}
\centering
\begin{tabularx}{\textwidth}{|Y|Y|Y|Y|Y|}
\hline
\textbf{Feature/Criteria} & \textbf{Demand Response} & \textbf{Microgrid Management} & 
\textbf{Renewable Energy Integration}
& \textbf{HVAC Control} \\
\hline
\textbf{Key Objectives} & Optimize energy usage and cost in response to utility signals, enhancing grid stability & Enhance grid resilience and efficiency, optimizing energy distribution and usage & Seamlessly integrate renewable energy into power systems, maximizing utilization while ensuring grid stability & Optimize HVAC systems for energy efficiency without compromising occupant comfort \\
\hline
\textbf{Challenges Addressed} & Adapting to dynamic pricing and demand, improving energy consumption efficiency & Managing diverse energy sources, ensuring reliable and efficient operation & Addressing variability and unpredictability of renewable sources & Balancing energy savings with thermal comfort requirements \\
\hline
\textbf{RL Approaches} & PPO \cite{jang2021using, azuatalam2020reinforcement}, MARL \cite{ahrarinouri2020multiagent, zhang2021testbed}, DQN \cite{lu2021deep}, MADDPG \cite{lu2020multi}

& DQN \cite{shojaeighadikolaei2021weather}, PPO \cite{zhang2023deep}, A3C \cite{nakabi2021deep},  
& MA-DRL \cite{cao2020reinforcement}, Q-learning \cite{ahrarinouri2020multiagent} 
& PPO \cite{azuatalam2020reinforcement}, Batch Constrained Munchausen Deep Q-learning \cite{liu2022safe}, Q-learning \cite{yuan2021study}, A3C \cite{zhong2022end}\\
\hline
\textbf{Methodology Highlights} & Meta-learning for simulation-experiment gap, multi-agent systems for residential energy management & Expert knowledge integration, operational flexibility with proximal policy optimization & Analysis on RL's role, multi-task learning for system-wide optimization & Architecture optimization for demand response, safe control strategies, and energy consumption reduction \\
\hline
\textbf{Outcomes} & Up to 22\% energy savings, efficient electricity usage management & Improved energy distribution and cost efficiency, increased resilience & Enhanced management of complex energy flows, significant performance improvements & Reduction in HVAC energy consumption, improved operational efficiency \\
\hline
\textbf{Future Directions} & Advanced cooperative strategies, bridging the simulation-experiment gap & Enhanced learning efficiency, integration with smart grid technologies & Scalability and adaptability of RL methods, robustness against environmental changes & Adaptability to diverse buildings, pre-training models, transfer learning applications \\
\hline
\textbf{Representative Studies} & 

Azuatalam et al. \cite{azuatalam2020reinforcement}, Jang et al. \cite{jang2021using}, Ahrarinouri et al. \cite{ahrarinouri2020multiagent}, Lu et al. \cite{lu2021deep}, Lu et al. \cite{lu2020multi}, Zhang et al. \cite{zhang2021testbed}

& 
Nakabi and Toivanen \cite{nakabi2021deep}, Hu and Kwasinski \cite{hu2021energy}, Zhang et al. \cite{zhang2023deep}, Zhang et al. \cite{zhang2022energy}, Shojaeighadikolaei et al. \cite{shojaeighadikolaei2021weather}, Du and Li \cite{du2019intelligent}

& 

Yang et al. \cite{yang2020reinforcement}, Cao et al. \cite{cao2020reinforcement}, Chen et al. (\cite{chen2022reinforcement}, Sivamayil et al. \cite{sivamayil2023systematic}, Perera and Kamalaruban \cite{perera2021applications}, Ahrarinouri et al. \cite{ahrarinouri2020multiagent}

& 

Azuatalam et al. \cite{azuatalam2020reinforcement}, Zhong et al. \cite{zhong2022end}, Sierla et al. \cite{sierla2022review}, Liu et al. \cite{liu2022safe}, Yuan et al. \cite{yuan2021study}, Biemann et al. \cite{biemann2021experimental}

\\
\hline
\end{tabularx}
\caption{Comparison of RL Approaches for Optimization in Energy Systems}
\label{tab:energy}
\end{table*}

\newcolumntype{Y}{>{\centering\arraybackslash}X}

\renewcommand{\arraystretch}{1} 
\renewcommand\tabularxcolumn[1]{m{#1}} 

\begin{table*}
\centering
\begin{tabularx}{\textwidth}{|Y|Y|Y|Y|Y|}
\hline
\textbf{Feature/Criteria} & \textbf{Motion Planning} & \textbf{Manipulation} & \textbf{Multi-robot Coordination} & \textbf{Human-robot Collab} \\
\hline
\textbf{Key Objectives} & Enable robots to navigate and perform tasks in dynamic environments & Enhance robotic interaction with objects and environments & Optimize collaborative actions of multiple robots for a common goal & Facilitate effective interaction and cooperation between humans and robots \\
\hline
\textbf{Challenges Addressed} & Navigating complex and dynamic environments, learning from interaction & Adapting to diverse objects, leveraging complex sensor inputs & Resource competition, obstacle avoidance in cooperative tasks & Adaptation to human behaviors, ensuring safety and making intelligent decisions \\
\hline
\textbf{RL Approaches} & PPO \cite{zhou2021robotic}, Q-learning \cite{yu2022reinforcement}, Soft Actor-Critic (SAC) \cite{cao2023reinforcement} 
& DDPG \cite{schuck2022dext}, Double DQN \cite{joshi2020robotic} 
& Multi-Robot Coordination with Deep Reinforcement Learning (MRCDRL) \cite{wang2020mrcdrl}, Multi Agent Deep Reinforcement Learning \cite{lan2021towards}
& DQN \cite{ghadirzadeh2020human, iucci2021explainable}, SAC \cite{shafti2020real, thumm2023human}, DDPG \cite{el2020towards}, Double DQN \cite{cai2022framework} \\
\hline
\textbf{Methodology Highlights} & EfficientLPT \cite{cao2023reinforcement} for space robots, curriculum learning for robotic arms & Visuo-motor feedback, dexterous grasping in sparse environments & MRCDRL \cite{wang2020mrcdrl} for cooperative action, MARL for pick-and-place optimization & Human-centered DRL, explainable RL for interaction quality enhancement \\
\hline
\textbf{Outcomes} & Improved planning accuracy, learning from human demonstrations & Significant outperformance in grasping tasks, adaptability to grippers & Effective resource allocation and dynamic obstacle avoidance, applicability in smart manufacturing & Enhanced coordination in packaging tasks, adaptability to user habits during collaboration \\
\hline
\textbf{Future Directions} & Integration with sensory feedback, real-time adaptation & Incorporation of more complex sensory modalities, tactile feedback & Scalable coordination strategies for larger teams, integration with smart environments & Personalized adaptation to human habits, enhancing safety and interpretability \\
\hline
\textbf{Representative Studies} & 
Wang et al. \cite{wang2021survey}, Cao et al. \cite{cao2023reinforcement}, Zhou et al. \cite{zhou2021robotic}, Yu and Chang \cite{yu2022reinforcement}

& 
Joshi et al. \cite{joshi2020robotic}, Schuck et al. \cite{schuck2022dext}, Han et al. \cite{han2023survey}, Rivera et al. \cite{rivera2021reward}, Beigomi and Zhu \cite{beigomi2024enhancing}

& 
Wang and Deng \cite{wang2020mrcdrl}, Lan et al. \cite{lan2021towards}, Yang \cite{yang2021reinforcement}, Lan et al. \cite{lan2022coordination}, Sadhu and Konar \cite{sadhu2020multi}, Khamassi \cite{khamassi2020adaptive}

& 

Ghadirzadeh et al. \cite{ghadirzadeh2020human}, Iucci et al. \cite{iucci2021explainable}, Shafti et al. \cite{shafti2020real}, Cai et al. \cite{cai2022framework}, Thumm et al. \cite{thumm2023human}, El-Shamouty et al. \cite{el2020towards}

\\
\hline
\end{tabularx}
\caption{Comparison of RL Approaches for Optimization in Robotics}
\label{tab:robotics}
\end{table*}

\subsection{Robotics}
RL is revolutionizing robotics, making significant strides across motion planning, grasping and manipulation, multi-robot coordination, and human-robot collaboration, thereby addressing intricate challenges inherent in the field. In motion planning, RL, particularly DRL and innovative methodologies like curriculum learning, empowers robots to adeptly navigate and execute tasks in dynamic environments, enhancing adaptability and task performance \cite{wang2021survey, cao2023reinforcement, zhou2021robotic, yu2022reinforcement}. Grasping and manipulation benefit from DRL's ability to process complex sensor inputs, enabling robots to interact with diverse objects and environments with unprecedented flexibility and efficiency \cite{joshi2020robotic, schuck2022dext, han2023survey, rivera2021reward, beigomi2024enhancing}. Multi-robot coordination leverages DRL and MARL to facilitate sophisticated collaborative strategies among robots, optimizing collective actions to achieve common goals in complex and dynamic tasks \cite{wang2020mrcdrl, lan2021towards, yang2021reinforcement, lan2022coordination, sadhu2020multi, khamassi2020adaptive}. Human-robot collaboration (HRC) sees advancements through DRL's capacity for learning from interactions and adapting to human behaviors, significantly improving cooperation in tasks ranging from manufacturing to daily assistance \cite{ghadirzadeh2020human, iucci2021explainable, shafti2020real, cai2022framework, thumm2023human, el2020towards}. Future research directions emphasize the integration of sensory feedback for real-time adaptation in motion planning, enhancing grasping tasks with complex sensory and tactile feedback, developing scalable coordination strategies for larger robot teams, and personalizing HRC to adapt to human habits while enhancing safety and interpretability. Table \ref{tab:robotics} succinctly encapsulates these domains by detailing key objectives, challenges addressed, RL approaches, methodology highlights, outcomes, and future directions, alongside representative studies demonstrating RL's transformative impact on robotics.

\section{CHALLENGES, STATE OF THE ART, AND FUTURE DIRECTIONS}
There has been a significant progress in the field of RL for optimization in automation; however, there are still challenges to be addressed. Table \ref{tab:challenges} gives a comparison of these challenges, along with the state of the art in the field and future directions that we will discuss in this section.

\newcolumntype{Y}{>{\centering\arraybackslash}X}

\renewcommand{\arraystretch}{1} 
\renewcommand\tabularxcolumn[1]{m{#1}} 

\begin{table*}
\centering
\begin{tabularx}{\textwidth}{|Y|Y|Y|Y|Y|Y|}
\hline
\textbf{Challenges} & \textbf{Description} & \textbf{RL Approaches} & \textbf{State of the Art} & \textbf{Future Directions} & \textbf{Related Studies} \\
\hline
\textbf{Sample Efficiency and Scalability} & Reducing the data needed for learning and ensuring scalability & PPO, SAC, Model-based Policy Optimization (MBPO), Dreamer,
IMPALA, Acme & Model-based RL with planning algorithms, off-policy learning with prioritized experience replay, and large-scale distributed RL systems & Focus on algorithms with adaptive learning rates and cross-domain transfer learning & Tianyue Cao \cite{cao2020study}, Florian E. Dorner \cite{dorner2021measuring}, Suri et al. \cite{suri2020maximum} Yang et al. \cite{yang2020sample}, Ball et al. \cite{ball2023efficient} Li et al. \cite{ li2020breaking}, Ly et al. \cite{ly2023elastic}, Wang et al. \cite{wang2022sample} \\
\hline
\textbf{Safety and Robustness} & Ensuring RL policies perform safely under uncertain conditions & Constrained Policy Optimization (CPO), Lyapunov-based approaches, State-wise Safe RL, Probabilistic constraint methods & Formal methods for policy verification, robust adversarial training, and safe exploration techniques & Integrating formal verification methods and enhancing human-RL interaction & Hao Xiong and Xiumin Diao \cite{xiong2021safety}, Li et al. \cite{li2023safe}, Liu et al.\cite{liu2022robustness}, Emam et al. \cite{emam2022safe}, Li et al. \cite{li2021safe}, \cite{zanon2020safe}, Queeney et al. \cite{queeney2023optimal}, Md Asifur Rahman and Sarra M. Alqahtani \cite{rahman2023task}, Wang et al. \cite{wang2020falsification}  \\
\hline
\textbf{Interpretability and Trustworthiness} & Developing RL models whose actions are transparent and understandable & 

MARL, Q-learning, Deep RL, DQN, PPO, TD3, SAC

& Feature attribution, policy distillation, and interpretable models like decision trees and attention mechanisms & Improving the foundation of interpretable models and applying self-supervised learning for interpretable representations & Glanois et al. \cite{glanois2021survey}, Duo Xu and Faramarz Fekri \cite{xu2021interpretable}), Mansour et al. \cite{mansour2022there}, Eckstein et al. \cite{eckstein2021reinforcement}, Alharin et al. \cite{alharin2020reinforcement}, Shi et al. \cite{shi2020self}, Dao et al. \cite{dao2021learning}. \\
\hline
\textbf{Transfer Learning and Meta-learning} & Enabling RL systems to rapidly adapt to new tasks using knowledge from past experiences & 
A3C, Meta-RL, Meta-RL with Context-conditioned Action Translator (MCAT), TD3

& Context-based meta-learning frameworks, multi-task learning techniques, and fine-tuning pre-trained models & Developing algorithms that generalize across a wider range of tasks and enhance transfer learning capabilities & Hospedales et al. \cite{hospedales2021meta}, Guo et al. \cite{guo2022learning}, Narvekar et al. \cite{narvekar2020curriculum}, Varma et al. \cite{varma2022effective}, Sasso et al. \cite{sasso2021multi}, Ren et al. \cite{ren2022efficient} \\
\hline
\textbf{Real-world Deployment and Integration} & Bridging the gap between theoretical advancements and practical utility in RL deployment & 

Behavior-Regularized Model-ENsemble (BREMEN), Distributional Maximum a Posteriori Policy Optimization (DMPO), 
Distributed Distributional Deterministic Policy Gradient (D4PG)

& Scalable RL architectures, robust policy deployment strategies, and open-source benchmarks and toolkits & Prioritizing deployment efficiency, enhancing human-RL interaction, and fostering academia-industry collaboration & Dulac-Arnold et al. \cite{dulac2020empirical}, Matsushima et al. \cite{matsushima2020deployment}, Yahmed et al. \cite{yahmed2023deploying}, Li et al. \cite{li2023deploying}, Garau-Luis et al. \cite{garau2021evaluating}, Kanso and Patra \cite{kanso2022engineering} \\
\hline
\end{tabularx}
\caption{Current Challenges, State of the Art, and Future Directions for RL for Optimization in Automation}
\label{tab:challenges}
\end{table*}

\subsection{Sample Efficiency and Scalability}
Sample efficiency and scalability are vital in RL to minimize training data and ensure solutions scale with task complexity. These challenges are particularly important in real-world applications where data collection is expensive or time-consuming \cite{cao2020study, dorner2021measuring}.

Current efforts to enhance sample efficiency and scalability include making past samples more reflective of the current model \cite{cao2020study, dorner2021measuring}, using evolution strategies and efficient memory in experience replay \cite{suri2020maximum, yang2020sample}, incorporating offline data for online learning \cite{ball2023efficient, li2020breaking}, and leveraging adaptive learning techniques \cite{ly2023elastic, wang2022sample}.


Future research should aim at algorithms with adaptive learning rates, domain-specific knowledge integration, efficient computational resource use, and cross-domain transfer learning to further improve sample efficiency and scalability in RL applications.

\subsection{Safety and Robustness}
Ensuring safety and robustness in RL is crucial, especially for applications in critical domains like autonomous driving and healthcare. Safe RL algorithms aim to learn policies that satisfy safety constraints during both training and deployment \cite{garcia2015comprehensive}. 

Current strategies for ensuring safety include developing concepts of safety robustness \cite{xiong2021safety}, frameworks for robust policies \cite{li2023safe}, tackling observational adversarial attacks \cite{liu2022robustness}, integrating robust-control-barrier-function layers \cite{emam2022safe}, managing safety requirements with robust action governor \cite{li2021safe}, enforcing safety via robust Model Predictive Control (MPC) \cite{zanon2020safe}, offering robustness guarantees \cite{queeney2023optimal, rahman2023task}, improving policy robustness through falsification-based adversarial learning \cite{wang2020falsification}, and inducing a safety curriculum \cite{turchetta2020safe}. 


Future research directions should focus on developing scalable safe RL algorithms for high-dimensional continuous control tasks, integrating formal verification methods with RL, and improving the adaptability of safe RL algorithms to dynamic environments.

\subsection{Interpretability and Trustworthiness}
Ensuring RL models are interpretable and trustworthy is essential for applications in healthcare, autonomous systems, and finance, requiring transparent, understandable, and reliable decision-making processes.

Current research to improve interpretability includes distinguishing between interpretability and explainability \cite{glanois2021survey}, integrating symbolic logic with deep RL for transparency \cite{xu2021interpretable}, achieving policy interpretability in structured environments \cite{mansour2022there}, interpreting RL modeling in cognitive sciences \cite{eckstein2021reinforcement}, discovering interpretable features in vision-based RL \cite{shi2020self}, and introducing sparse evidence collection for human interpretation \cite{dao2021learning}.

Advancements will focus on foundational improvements to make models intrinsically understandable, incorporating human feedback, advancing feature discovery techniques, and applying self-supervised learning for natural interpretability, aiming for a deeper human understanding of RL behaviors.

\subsection{Transfer Learning and Meta-learning}
Transfer learning and meta-learning address the need for RL systems to efficiently adapt to new tasks using knowledge from past experiences, aiming to improve learning efficiency and generalization across various environments.

Hospedales et al. \cite{hospedales2021meta} highlight meta-learning's role in adaptability across tasks. Guo et al. \cite{guo2022learning} develop an action translator for meta-RL to enhance exploration and efficiency. Narvekar et al. \cite{narvekar2020curriculum} present a curriculum learning framework that uses task sequencing for improved learning in complex scenarios. Varma et al. \cite{varma2022effective} demonstrate the benefits of using pre-trained models like ResNet50 to boost RL performance. Sasso et al. \cite{sasso2021multi} and Ren et al. \cite{ren2022efficient} investigate multi-source transfer learning and meta-RL for fast adaptation based on human preferences.

Future efforts will focus on algorithms that better generalize across diverse tasks, with a push towards unsupervised and self-supervised learning to advance transfer learning capabilities. There's also a growing interest in models that autonomously leverage past knowledge.

\subsection{Real-world Deployment and Integration}
Real-world RL model deployment involves overcoming the divide between theoretical research and practical application, ensuring model robustness, and aligning simulated training environments with real-world conditions.

Dulac-Arnold et al. \cite{dulac2020empirical} highlight real-world RL deployment challenges, introducing benchmarks for complexity. Matsushima et al. \cite{matsushima2020deployment} focus on efficient deployment with minimal data. Yahmed et al. \cite{yahmed2023deploying} outline deployment challenges and emphasize the need for solutions. Li et al. \cite{li2023deploying} advocate for incorporating human feedback during deployment for safety. Garau-Luis et al. \cite{garau2021evaluating} discuss DRL deployment advancements, while Kanso and Patra \cite{kanso2022engineering} discuss engineering solutions for RL scalability.

Future efforts will center on algorithms and frameworks enhancing deployment efficiency and real-world relevance, generalization from simulations to reality, improving human-RL interactions, and robust, scalable deployment platforms. Domain-specific challenges and academia-industry collaboration are pivotal for RL's real-world success.

\section{CONCLUSION}

Reinforcement Learning (RL) has showcased its vast capabilities in sectors such as manufacturing, energy systems, and  robotics, driven by deep learning innovations that tackle complex challenges. Despite these advancements, real-world deployment introduces challenges requiring extensive research for practical RL implementation. This review emphasizes the need for improved sample efficiency, model safety, interpretability, and real-world integration strategies. To meet these requirements, a comprehensive approach is necessary, integrating algorithmic advancements, domain-specific insights, robust benchmarks, and understanding the balance between theory and practice. Moreover, integrating human feedback and ethical considerations is crucial for the responsible deployment of RL.  Ultimately, RL's transition from theory to a key AI component marks significant progress, with ongoing efforts expected to overcome current obstacles, leveraging RL's full potential in intelligent decision-making and system optimization.

\bibliographystyle{IEEEtran} 
\bibliography{root.bib}

\end{document}